\documentclass[11pt,a4paper,svgnames,x11names]{article}
\usepackage[hyperref]{emnlp2020}
\usepackage{times}
\usepackage{latexsym}

\usepackage{amsmath}
\usepackage{amssymb}
\usepackage{graphicx}
\usepackage{verbatim}
\usepackage{tabularx}
\usepackage{booktabs}
\usepackage{cleveref}

\DeclareMathOperator*{\argmax}{arg\,max}

\usepackage{microtype}

\aclfinalcopy %

\title{{\it CopyNext:} Explicit Span Copying and Alignment\\in Sequence to Sequence Models}
\author{{Abhinav Singh\textsuperscript{1,2} ~~ Patrick Xia\textsuperscript{1} ~~ Guanghui Qin\textsuperscript{1}} \\ \textbf{Mahsa Yarmohammadi\textsuperscript{1} ~~ Benjamin Van Durme\textsuperscript{1}}
	\\
	\textsuperscript{1}Johns Hopkins University  ~~ \textsuperscript{2}Bloomberg L.P.   \\
	
	\texttt{{abhinavsingh282@gmail.com}, \texttt{\{paxia,vandurme\}@cs.jhu.edu},} \\ 
	\texttt{{\{qin,mahsa\}@jhu.edu}} 
	
}

\date{}

\begin{document}
\maketitle
\begin{abstract}
Copy mechanisms are employed in sequence to sequence models (seq2seq) to generate reproductions of words from the input to the output.  These frameworks, operating at the lexical \textit{type} level, fail to provide an explicit alignment that records where each \textit{token} was copied from. Further, they require contiguous token sequences from the input (spans) to be copied individually. We present a model with an explicit token-level copy operation and extend it to copying entire spans. %
Our model provides hard alignments between \textit{spans} in the input and output, allowing for nontraditional applications of seq2seq, like information extraction. %
We demonstrate the approach on Nested Named Entity Recognition, achieving near state-of-the-art accuracy with an order of magnitude increase in decoding speed.
\footnote{Our source code:  \url{https://github.com/abhinonymous/copynext}}
\end{abstract}

\section{Introduction}
Sequence transduction converts a sequence of input tokens to a sequence of output tokens. It is a dominant framework for generation tasks, such as machine translation, dialogue, and summarization. Seq2seq can also be used for Information Extraction (IE), where the target structure is \emph{decoded} as a linear output based on an encoded (linear) representation of the input.

As IE is traditionally considered a structured prediction task, it remains today that IE systems are assumed to produce an annotation on the input text.  That is, predicting which specific tokens of an input string led to, e.g., the label of {\sc Person}.  This is in contrast to text generation which rarely, if ever, needs \emph{hard alignments} between the input and the desired output.  Our work explores a novel extension to seq2seq that provides such alignments.

\begin{figure}
\centering
\includegraphics[width=7.7cm]{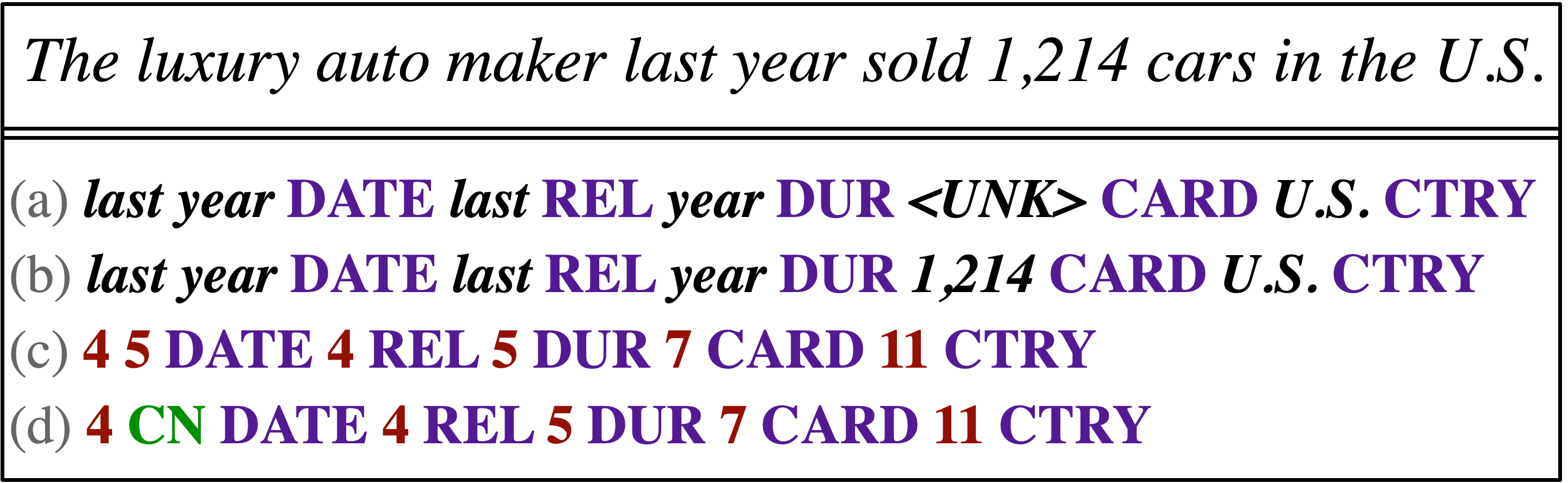}
\caption{Sequence transduction outputs for nested named entities in an example sentence using: (a) seq2seq, (b) pointer network, (c) Copy-only, and (d) CopyNext model. The numbers are predictions of \textit{indices} corresponding to the tokens in the input sequence. CN refers to the CopyNext symbol, our proposed method of denoting the operation that copies the next token from the input. In (d), the next token from token 4 would be 5.}
\label{fig:example}
\end{figure}

Specifically, we extend pointer (or copy) networks. Unlike the algorithmic tasks originally targeted by \citet{vinyals2015pointer}, tasks in NLP tend to copy {\em spans} from the input rather than discontiguous tokens. This is prevalent for copying named entities in dialogue \cite{gu-etal-2016-incorporating,eric-manning-2017-copy}, entire sentences in summarization \cite{see-etal-2017-get,song-etal-2018-structure}, or even single words (if subtokenized). The need to efficiently copy spans motivates our introduction of an inductive bias that copies contiguous tokens. Like a pointer network, our model copies the first token of a span. However, for subsequent timesteps, our model generates a ``CopyNext'' symbol (CN) instead of copying another token from source. CopyNext represents the operation of copying the word following the last predicted word from the input sequence. \autoref{fig:example} highlights the difference between output sequences for several transductive models, including our CopyNext model. %

\begin{figure*}
\centering
\includegraphics[width=\textwidth]{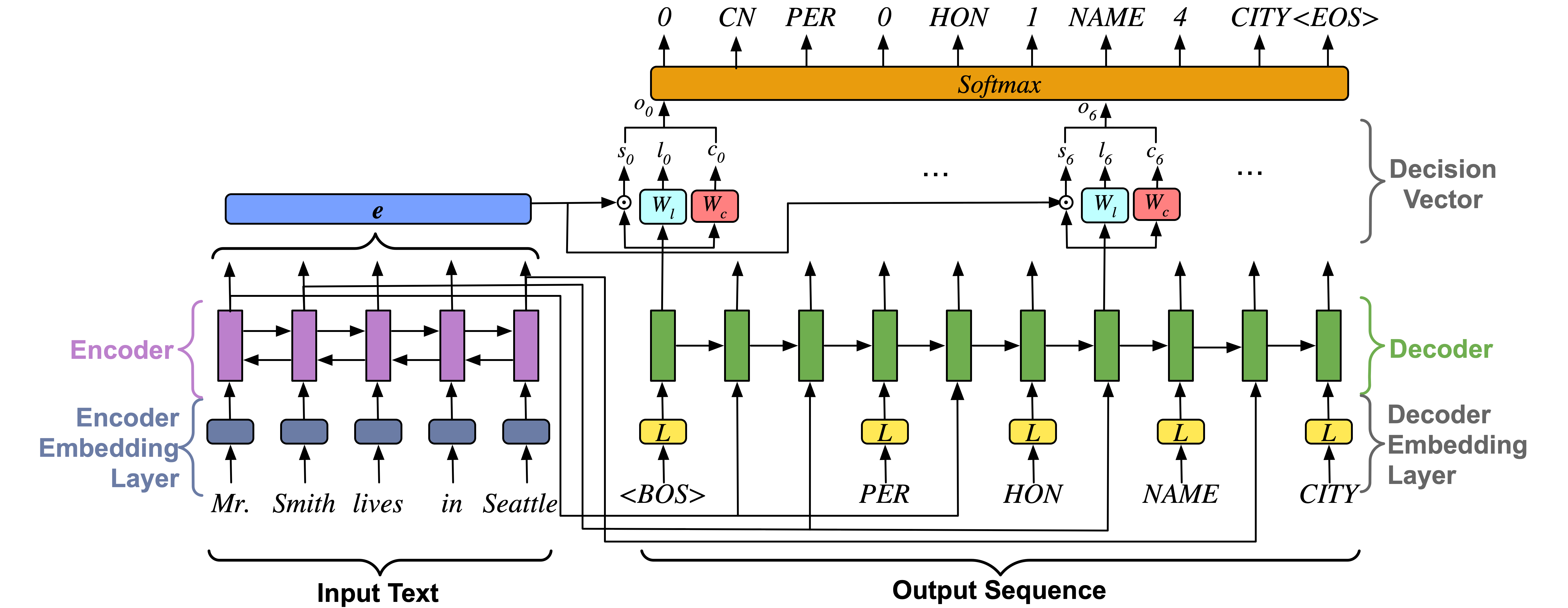}
\caption{For each decoder timestep a decision vector chooses between labeling, a CopyNext operation, or pointing to an input token. The decoder input comes from either an encoder state or a label embedding.}
\label{fig:model}
\end{figure*}

We apply our model for the Nested Named Entity Recognition (NNER) task  \citep{ringland-etal-2019-nne}. %
Unlike traditional named entity recognition, named entity mentions in NNER may be subsequences of other named entity mentions (such as \textit{[[last] year]} in \autoref{fig:example}). We find that both explicit copying and CopyNext lead to a system faster than prior work and better than a simple seq2seq baseline. It is, however, outperformed by a much slower model that performs an exhaustive search over the space of potential labels, a solution that does not scale to large complex label sets.

\section{Related Work}
Pointer networks \cite{vinyals2015pointer,jia2016data,merity2016pointer} are seq2seq models that employ a soft attention distribution \cite{bahdanau2014neural} to produce an output sequence consisting of values from the input sequence.
Pointer-generator networks \cite[\textit{inter alia}]{miao2016language,gulcehre-etal-2016-pointing} extend the range of output types by combining the distribution from the pointer with a vocabulary distribution from a generator. Thus, these models operate on the \textit{type} level.
In contrast, our model operates at the \textit{token} level. Instead of using soft attention distribution of the encoder states, we use hard attention, resulting in a single encoder state, or a single token, to feed to the decoder. This enables explicit copying of span offsets. 

Closest to our work, \citet{zhou2018sequential} and \citet{panthaplackel2020copy} have tackled span copying by extending pointer-generator networks and predicting both start and end indices of entire spans that need to be copied. Using those offsets, they perform a forced decoding of the predicted tokens within the span. These works focus on text generation tasks, like sentence summarization, question generation, and editing. %
In contrast, we are concerned with information extraction tasks as transduction, where hard alignments to the input sentence are crucial and output sequences must represent a valid linearized structure. Specifically, we study nested named entity recognition (NNER).

Prior work uses several approaches to model NNER: machine reading comprehension \cite{li2019unified}, transition-based methods \cite{wang-etal-2018-neural-transition}, mention hypergraphs \cite{lu2015joint,wang-lu-2018-neural,katiyar2018nested}, and seq2seq models \cite{strakova2019neural}.

\section{Model Description}

We formulate the task as transforming the input sentence $X$ to a linearized sequence $Y$ which represents the gold structure: labeled spans. Specifically, $Y$ contains input word indices, CopyNext symbols, and labels from a label set $\mathcal{L}$. %

As described earlier, the model (Figure \ref{fig:model}) is reminiscent of pointer networks. %
We extend its capabilities by introducing the notion of a ``Copy Next'' operation where the network predicts to copy the word sequentially after the previous prediction.

\subsection{Encoder}

\paragraph{Embedding Layer}
This layer embeds a sequence of tokens $X = \langle x_{1},x_{2},...,x_{N^\prime} \rangle$ into a sequence of vectors $\mathbf{x} = \langle \mathbf{x}_1,\mathbf{x}_2,...,\mathbf{x}_N \rangle$ by using (possibly contextualized) word embeddings. %
The gold labels are adjusted to account for tokenization.

\paragraph{Architecture}
The input embedding is further encoded by a stacked bidirectional LSTM \citep{hochreiter1997long} into encoder states $ \mathbf{e} = \langle \mathbf{e}_1,\mathbf{e}_2,...,\mathbf{e}_N \rangle$ where each state is a concatenation of the forward ($\overrightarrow{f}$) and backward ($\overleftarrow{f}$) outputs of the last layer of the LSTM and $\mathbf{e}_i \in \mathbb{R}^{D}$:
\begin{align}
\mathbf{e}_i^j = [\overrightarrow{f}^j(\mathbf{e}_i^{j-1}, \mathbf{e}_{i-1}^j); \overleftarrow{f}^j(\mathbf{e}_i^{j-1}, \mathbf{e}_{i+1}^j)],
\end{align}
where $\mathbf{e}_i^j$ is the $j$-th layer encoder hidden state at timestep $i$ and $D$ is the hidden size of the LSTM.

\subsection{Decoder}
The target for the transducer is the linearized representation of the nested named entity spans and labels.
We generate a decision $y$ that either points to
(a) a timestep in the encoder sequence, marking the starting index of a span, or
(b) the CopyNext symbol, which operates by advancing the right boundary of the span to include the next (sub)word of the input sequence, or 
(c) a label $l \in \mathcal{L}$, signifying both the end of the span and classifying the span.

\paragraph{Input Embeddings} %
We learn $D$-dimensional embeddings for each label $l \in \mathcal{L}$. 
The vectors corresponding to the start index of a span and the CopyNext operation are the encoder outputs $\mathbf{e}_i$ where $i$ is equal to the start index or index pointed to by CopyNext 
and are fed directly to the decoder.\footnote{We will use $\mathbf{e}_i$ to refer to $\mathbf{e}_i^{(-1)}$.}

\paragraph{Architecture}
The decoder is a stacked LSTM taking as input either 
an encoder state $\mathbf{e}_i$  or a label embedding %
and produces decoder state $\mathbf{d}_t \in \mathbb{R}^{D}$.

\paragraph{Decision Vector}
We predict scores for making a labeling decision, a CopyNext operation, or pointing to a token in the input. 
At each decoding step $t$, 
for labels, we train a linear layer $W_L \in \mathbb{R}^{D \times |\mathcal{L}|}$ with input $\mathbf{d}_t$ and output scores $\mathbf{l}_t$. Likewise, we do the same for the CopyNext symbol using a linear layer $W_C \in \mathbb{R}^{D \times 1}$ with input $\mathbf{d}_t$ and output score $\mathbf{c}_t$. 
The score of pointing to an index $i$ in the input sequence is calculated by dot product: 
$s_{t}^i = \mathbf{e}_i \cdot \mathbf{d}_t$.
The decision distribution $\mathbf{y}_t$ is then:
\begin{equation}
    \mathbf{y}_t = \text{softmax}([\mathbf{s}_t;\mathbf{l}_t;\mathbf{c}_t]),\ \mathbf{y}_t \in \mathbb{R}^{N + |\mathcal{L}| + 1}.
\end{equation}

\subsection{Training and Prediction}

Our training objective is the cross-entropy loss:
\begin{equation}
    \ell = \sum_{t}\sum_{k} \delta_{y_{t}^k,y_t^\star}\log(y_{t}^k)
\end{equation}
where $y^\star$ is the gold decision, $k \in [0,{N + |\mathcal{L}| + 1})$ (representing all three kinds of possible decisions: index, label or CopyNext)
and $\delta_{y_{t}^k,y_t^\star}$ is 1 if $y_t^k = y_t^\star$ and 0 otherwise. 
The summation over index $t$ covers the whole dataset.

At prediction time we find the decision $\overline{y}_t$ with the greatest probability ($\overline{y}_t =\argmax_i(y_t^i)$) at decoder step $t$.\footnote{Initial experiments with beam search suggest an expensive tradeoff between time and performance (Appendix \ref{sec:app_expr}).}
The input to the decoder at $t+1$ timestamp can be one of three things:
(1) the output $\mathbf{e}_i$ of the encoder when $\overline{y}_t$ points to the index $i$ of the input sequence, (2) the embedding of the label $l$ predicted at $t$ when $\overline{y}_t$ points to the label $l\in\mathcal{L}$,
or (3) the output $\mathbf{e}_{i+1}$ of the encoder where $i$ was the input to the decoder at $t$ when $\overline{y}_t$ points to the CopyNext operation.
The decoder halts when the $\langle \mathit{EOS} \rangle $ label is predicted or the maximum output sequence length is reached.

\begin{figure}[t]
\centering
\includegraphics[width=7cm]{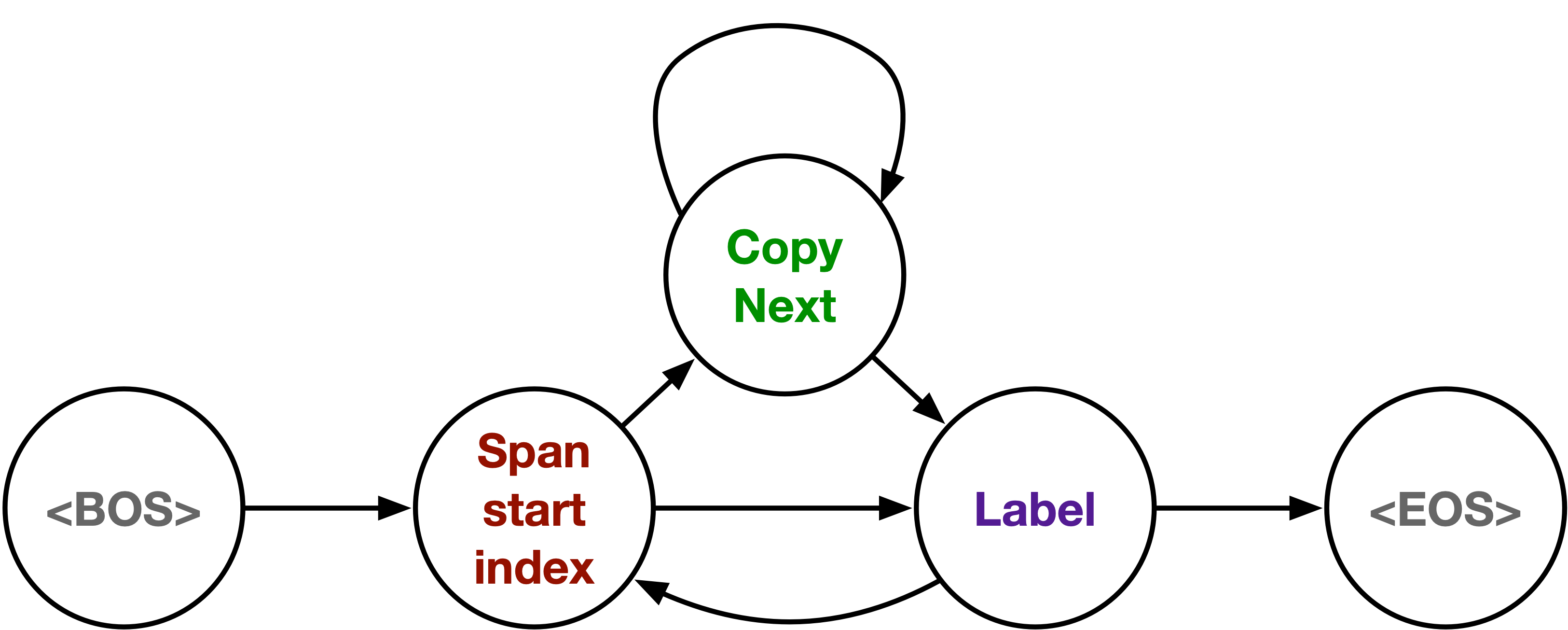}
\caption{State machine of a well-formed predicted sequence for masking of the decision vector at inference.}
\label{fig:StateMachine}
\end{figure}
To ensure well-formed target output sequences, we use a state machine (\autoref{fig:StateMachine}) to mask parts of $\mathbf{\overline{y}}_t$ that would lead to an illegal sequence at $t+1$.  %

\begin{table}[t]
\small
\centering
    \begin{tabular}{lccc}
    \toprule
    \textbf{Embedding}  & \textbf{P}     & \textbf{R}     & \textbf{F1}    \\ \midrule
    RoBERTa       & \textbf{90.6} & \textbf{82.6} & \textbf{86.4}   \\ 
    BERT large cased     & 90.4 & 78.2 & 83.9  \\ 
    XLNet large cased    & 79.1 & 52.6 & 63.2  \\ 
    SpanBERT large cased & 89.1 & 80.2 & 84.4  \\ 
    ELMo                 & 82.1 & 73.3 & 77.4  \\ 
    GloVe                & 76.5 & 67.8 & 71.9  \\ 
    \bottomrule
\end{tabular}
\caption{A comparison of networks used for embedding input tokens before feeding them into the encoder LSTM. These are evaluated on NNER.}
\label{table:EmbeddingNetworks}
\end{table}

\section{Experiments and Results}

Our experiments analyze the effects of various choices in different components of the system. We use the NNE dataset and splits from \citet{ringland-etal-2019-nne}, resulting in 43,457, 1,989, and 3,762 sentences in the training, development, and test splits.
Experiments for model development and analysis use the development set.

\paragraph{Text Representation}
We first establish the best performing text embeddings which we fix for the rest of the experiments. The rationale is that the embedding will provide an orthogonal boost in accuracy to the network with respect to the other changes in the network structure. We find in \autoref{table:EmbeddingNetworks} that RoBERTa large \citep{roberta} is best.\footnote{We experimented with mean pooling RoBERTa vectors for subwords \citep{zhang-etal-2019-amr} to maintain the same span lengths as input. Pooling subword units led to poorer performance (see Appendix \ref{sec:app_expr} Table \ref{table:subwordpooling}).}

\begin{table}[t]
\small
\centering
\resizebox{\columnwidth}{!}{

\begin{tabular}{lcccc}
\toprule
 \textbf{Model}       & \textbf{P}     & \textbf{R}     & \textbf{F1}  &\textbf{Speed}  \\ \midrule
Hypergraph &  91.8 & 91.0 & 91.4& 1.0x\\
Transition & 77.4 & 70.1 & 73.6 & 6.3x\\
\midrule
Seq2seq & 86.6 & 63.6 & 75.4 & (*) \\
Copy &85.8	&81.3	 &83.5           & 16.7x \\
CopyNext    & 88.7	& 84.7	& 86.7     &16.7x \\
 \bottomrule
\end{tabular}
}
\caption{NNER accuracy and speed on the test set for external baselines and our models. *Seq2seq is based on a reference implementation to ensure correctness, but not efficiency: it has the same asymptotics as the Copy and CopyNext models, and can be considered similar in speed.}%
\label{table:results}
\end{table}

\paragraph{Linearization Strategy}
  Previous work \citep{zhang-etal-2019-amr} has shown that linearization scheme affects model performance. We experiment with several variants of sorting spans in ascending order based on their start index. We also try sorting based on end index and copying the \textit{previous} token instead. We find sorting based on end index performs poorly, while sorting by start all perform similarly. Our final linearization strategy sorts by start index, then span length (longer spans first). Additional ties (in span label) are broken randomly.

\paragraph{RoBERTa Embedding Layer}
Recent work suggests that NER information may be stored in the lower layers of an encoder \citep{hewitt-manning-2019-structural,tenney2019bert}. We found using the 15th layer of RoBERTa rather than the final one (24th), is slightly helpful (see Appendix \ref{sec:app_expr}).%

\paragraph{NNER Results}
In \autoref{table:results}, we evaluate our best-performing (dev.) model on the test set. We compare our approach against the previous best approaches reported in \citet{ringland-etal-2019-nne}: hypergraph-based \citep[Hypergraph,][]{wang-lu-2018-neural} and transition-based \citep[Transition,][]{wang-etal-2018-neural-transition} models proposed to recognize nested mentions. We also contrast the CopyNext model against a baseline seq2seq model and one with only a hard copy operation (see (a) and (c) in \autoref{fig:example}). %
Prior work \citep{wang-lu-2018-neural} has given an analysis of the run-time of their approach. Based on their concern about asymptotic speed we also provide the following analysis and practical speed efficiency of the systems and their accuracies.\footnote{Efficiency is measured using wall clock time for the entire test set, performed with Intel Xeon 2.10GHz CPU and a single GeForce GTX 1080 TI GPU.} 

We find that Hypergraph outperforms the CopyNext model by 4.7 F1 with most of the difference in recall. This is likely due to the exhaustive search used by Hypergraph, as our model is 16.7 times faster. %
An analysis of their code and algorithm reveals that their lower bound time complexity $\Omega(mn)$ is higher than ours $\Omega(n)$, $n$ is length of input sequence and $m$ is number of mention types. Since the average decoder length is low, the best case scenario often occurs. The Transition system has 6.3 times faster prediction speed compared to Hypergraph, however, it comes with 17.8\% absolute drop in F1 accuracy. Our model is substantially faster than both. Furthermore, we show that both an explicit Copy and the CopyNext operation are useful, resulting in gains of 8.1 F1 and 11.3 F1 over a seq2seq baseline.%

\begin{figure}
\centering
\includegraphics[width=7cm]{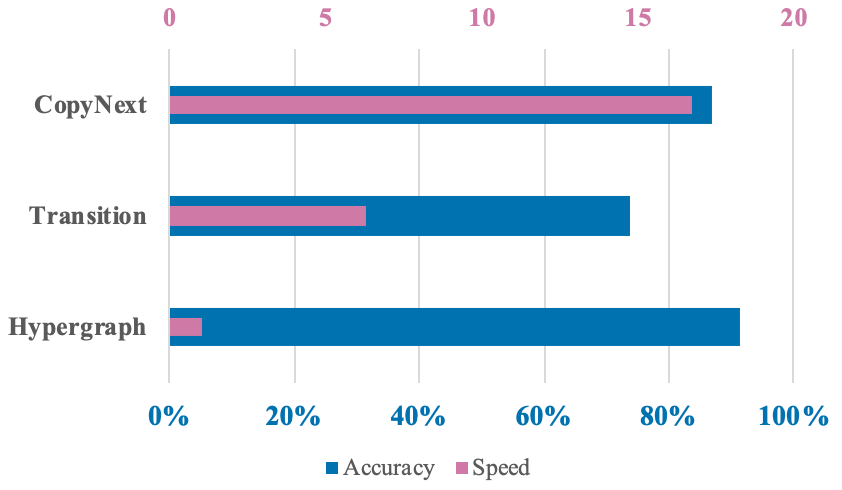}
\caption{Performance in terms of Accuracy (\%F1) and Speed (relative to Hypergraph). The CopyNext model is nearly as accurate as Hypergraph while over 16 times faster.}
\label{fig:accuracy-speed}
\end{figure}

\paragraph{NNER Error Analysis}
The errors made by the model on the development set can be clustered broadly into four main types: (1) correct span detection but mislabeled, (2) correct label but incorrect span detection (either subset or superset of correct span), (3) 
both span and label were incorrectly predicted, and (4) missing spans entirely. Table \ref{table:errors} in Appendix \ref{sec:app_expr} provides examples.

\section{Conclusion and Future Work}

We propose adopting pointer and copy networks with hard attention and extending these models with a CopyNext operation, enabling sequential copying of spans given just the start index of the span. On a traditionally structured prediction task of NNER, we use a sequence transduction model with the CopyNext operation, leading to a competitive model that provides a 16.7x speedup relative to current state of the art (which performs an exhaustive search), at a cost of 4.7\% loss in F1, largely due to lower recall.

Our model is a step forward in structured prediction as sequence transduction.  We have found in initial experiments on event extraction similar relative improvements to that discussed here: future work will investigate applications to richer transductive semantic parsing models \cite{zhang-etal-2019-amr, cai-lam-2019-core}.

\section*{Acknowledgments}        
This work was supported in part by IARPA BETTER (\#2019-19051600005), DARPA AIDA (FA8750-18-2-0015) and KAIROS (FA8750-19-2-0034). The views and conclusions contained in this work are those of the authors and should not be interpreted as necessarily representing the official policies, either expressed or implied, or endorsements of DARPA, ODNI, IARPA, or the U.S. Government. The U.S. Government is authorized to reproduce and distribute reprints for governmental purposes notwithstanding any copyright annotation therein.

\bibliography{anthology,bib}

\begin{thebibliography}{30}
\expandafter\ifx\csname natexlab\endcsname\relax\def\natexlab#1{#1}\fi

\bibitem[{Bahdanau et~al.(2014)Bahdanau, Cho, and Bengio}]{bahdanau2014neural}
Dzmitry Bahdanau, Kyunghyun Cho, and Yoshua Bengio. 2014.
\newblock Neural machine translation by jointly learning to align and
  translate.
\newblock \emph{arXiv preprint arXiv:1409.0473}.

\bibitem[{Cai and Lam(2019)}]{cai-lam-2019-core}
Deng Cai and Wai Lam. 2019.
\newblock \href {https://doi.org/10.18653/v1/D19-1393} {Core semantic first: A
  top-down approach for {AMR} parsing}.
\newblock In \emph{Proceedings of the 2019 Conference on Empirical Methods in
  Natural Language Processing and the 9th International Joint Conference on
  Natural Language Processing (EMNLP-IJCNLP)}, pages 3799--3809, Hong Kong,
  China. Association for Computational Linguistics.

\bibitem[{Devlin et~al.(2019)Devlin, Chang, Lee, and
  Toutanova}]{devlin-etal-2019-bert}
Jacob Devlin, Ming-Wei Chang, Kenton Lee, and Kristina Toutanova. 2019.
\newblock \href {https://doi.org/10.18653/v1/N19-1423} {{BERT}: Pre-training of
  deep bidirectional transformers for language understanding}.
\newblock In \emph{Proceedings of the 2019 Conference of the North {A}merican
  Chapter of the Association for Computational Linguistics: Human Language
  Technologies, Volume 1 (Long and Short Papers)}, pages 4171--4186,
  Minneapolis, Minnesota. Association for Computational Linguistics.

\bibitem[{Eric and Manning(2017)}]{eric-manning-2017-copy}
Mihail Eric and Christopher Manning. 2017.
\newblock \href {https://www.aclweb.org/anthology/E17-2075} {A copy-augmented
  sequence-to-sequence architecture gives good performance on task-oriented
  dialogue}.
\newblock In \emph{Proceedings of the 15th Conference of the {E}uropean Chapter
  of the Association for Computational Linguistics: Volume 2, Short Papers},
  pages 468--473, Valencia, Spain. Association for Computational Linguistics.

\bibitem[{Gu et~al.(2016)Gu, Lu, Li, and Li}]{gu-etal-2016-incorporating}
Jiatao Gu, Zhengdong Lu, Hang Li, and Victor~O.K. Li. 2016.
\newblock \href {https://doi.org/10.18653/v1/P16-1154} {Incorporating copying
  mechanism in sequence-to-sequence learning}.
\newblock In \emph{Proceedings of the 54th Annual Meeting of the Association
  for Computational Linguistics (Volume 1: Long Papers)}, pages 1631--1640,
  Berlin, Germany. Association for Computational Linguistics.

\bibitem[{Gulcehre et~al.(2016)Gulcehre, Ahn, Nallapati, Zhou, and
  Bengio}]{gulcehre-etal-2016-pointing}
Caglar Gulcehre, Sungjin Ahn, Ramesh Nallapati, Bowen Zhou, and Yoshua Bengio.
  2016.
\newblock \href {https://doi.org/10.18653/v1/P16-1014} {Pointing the unknown
  words}.
\newblock In \emph{Proceedings of the 54th Annual Meeting of the Association
  for Computational Linguistics (Volume 1: Long Papers)}, pages 140--149,
  Berlin, Germany. Association for Computational Linguistics.

\bibitem[{Hewitt and Manning(2019)}]{hewitt-manning-2019-structural}
John Hewitt and Christopher~D. Manning. 2019.
\newblock \href {https://doi.org/10.18653/v1/N19-1419} {{A} structural probe
  for finding syntax in word representations}.
\newblock In \emph{Proceedings of the 2019 Conference of the North {A}merican
  Chapter of the Association for Computational Linguistics: Human Language
  Technologies, Volume 1 (Long and Short Papers)}, pages 4129--4138,
  Minneapolis, Minnesota. Association for Computational Linguistics.

\bibitem[{Hochreiter and Schmidhuber(1997)}]{hochreiter1997long}
Sepp Hochreiter and J{\"u}rgen Schmidhuber. 1997.
\newblock Long short-term memory.
\newblock \emph{Neural computation}, 9(8):1735--1780.

\bibitem[{Jia and Liang(2016)}]{jia2016data}
Robin Jia and Percy Liang. 2016.
\newblock Data recombination for neural semantic parsing.
\newblock \emph{arXiv preprint arXiv:1606.03622}.

\bibitem[{Joshi et~al.(2020)Joshi, Chen, Liu, Weld, Zettlemoyer, and
  Levy}]{joshi2019spanbert}
Mandar Joshi, Danqi Chen, Yinhan Liu, Daniel Weld, Luke Zettlemoyer, and Omer
  Levy. 2020.
\newblock \href {https://transacl.org/index.php/tacl/article/view/1853}
  {Spanbert: Improving pre-training by representing and predicting spans}.
\newblock \emph{Transactions of the Association for Computational Linguistics},
  8(0):64--77.

\bibitem[{Katiyar and Cardie(2018)}]{katiyar2018nested}
Arzoo Katiyar and Claire Cardie. 2018.
\newblock Nested named entity recognition revisited.
\newblock In \emph{Proceedings of the 2018 Conference of the North American
  Chapter of the Association for Computational Linguistics: Human Language
  Technologies, Volume 1 (Long Papers)}, pages 861--871.

\bibitem[{Li et~al.(2019)Li, Feng, Meng, Han, Wu, and Li}]{li2019unified}
Xiaoya Li, Jingrong Feng, Yuxian Meng, Qinghong Han, Fei Wu, and Jiwei Li.
  2019.
\newblock A unified mrc framework for named entity recognition.
\newblock \emph{arXiv preprint arXiv:1910.11476}.

\bibitem[{Liu et~al.(2019)Liu, Ott, Goyal, Du, Joshi, Chen, Levy, Lewis,
  Zettlemoyer, and Stoyanov}]{roberta}
Yinhan Liu, Myle Ott, Naman Goyal, Jingfei Du, Mandar Joshi, Danqi Chen, Omer
  Levy, Mike Lewis, Luke Zettlemoyer, and Veselin Stoyanov. 2019.
\newblock \href {http://arxiv.org/abs/1907.11692} {Roberta: {A} robustly
  optimized {BERT} pretraining approach}.
\newblock \emph{CoRR}, abs/1907.11692.

\bibitem[{Lu and Roth(2015)}]{lu2015joint}
Wei Lu and Dan Roth. 2015.
\newblock Joint mention extraction and classification with mention hypergraphs.
\newblock In \emph{Proceedings of the 2015 Conference on Empirical Methods in
  Natural Language Processing}, pages 857--867.

\bibitem[{Merity et~al.(2016)Merity, Xiong, Bradbury, and
  Socher}]{merity2016pointer}
Stephen Merity, Caiming Xiong, James Bradbury, and Richard Socher. 2016.
\newblock Pointer sentinel mixture models.
\newblock \emph{arXiv preprint arXiv:1609.07843}.

\bibitem[{Miao and Blunsom(2016)}]{miao2016language}
Yishu Miao and Phil Blunsom. 2016.
\newblock Language as a latent variable: Discrete generative models for
  sentence compression.
\newblock In \emph{Proceedings of the 2016 Conference on Empirical Methods in
  Natural Language Processing}, pages 319--328.

\bibitem[{Panthaplackel et~al.(2020)Panthaplackel, Allamanis, and
  Brockschmidt}]{panthaplackel2020copy}
Sheena Panthaplackel, Miltiadis Allamanis, and Marc Brockschmidt. 2020.
\newblock \href {https://openreview.net/forum?id=SklM1xStPB} {Copy that!
  editing sequences by copying spans}.

\bibitem[{Pennington et~al.(2014)Pennington, Socher, and
  Manning}]{pennington2014glove}
Jeffrey Pennington, Richard Socher, and Christopher~D. Manning. 2014.
\newblock \href {http://www.aclweb.org/anthology/D14-1162} {Glove: Global
  vectors for word representation}.
\newblock In \emph{Empirical Methods in Natural Language Processing (EMNLP)},
  pages 1532--1543.

\bibitem[{Peters et~al.(2018)Peters, Neumann, Iyyer, Gardner, Clark, Lee, and
  Zettlemoyer}]{peters-etal-2018-deep}
Matthew Peters, Mark Neumann, Mohit Iyyer, Matt Gardner, Christopher Clark,
  Kenton Lee, and Luke Zettlemoyer. 2018.
\newblock \href {https://doi.org/10.18653/v1/N18-1202} {Deep contextualized
  word representations}.
\newblock In \emph{Proceedings of the 2018 Conference of the North {A}merican
  Chapter of the Association for Computational Linguistics: Human Language
  Technologies, Volume 1 (Long Papers)}, pages 2227--2237, New Orleans,
  Louisiana. Association for Computational Linguistics.

\bibitem[{Ringland et~al.(2019)Ringland, Dai, Hachey, Karimi, Paris, and
  Curran}]{ringland-etal-2019-nne}
Nicky Ringland, Xiang Dai, Ben Hachey, Sarvnaz Karimi, Cecile Paris, and
  James~R. Curran. 2019.
\newblock \href {https://doi.org/10.18653/v1/P19-1510} {{NNE}: A dataset for
  nested named entity recognition in {E}nglish newswire}.
\newblock In \emph{Proceedings of the 57th Annual Meeting of the Association
  for Computational Linguistics}, pages 5176--5181, Florence, Italy.
  Association for Computational Linguistics.

\bibitem[{See et~al.(2017)See, Liu, and Manning}]{see-etal-2017-get}
Abigail See, Peter~J. Liu, and Christopher~D. Manning. 2017.
\newblock \href {https://doi.org/10.18653/v1/P17-1099} {Get to the point:
  Summarization with pointer-generator networks}.
\newblock In \emph{Proceedings of the 55th Annual Meeting of the Association
  for Computational Linguistics (Volume 1: Long Papers)}, pages 1073--1083,
  Vancouver, Canada. Association for Computational Linguistics.

\bibitem[{Song et~al.(2018)Song, Zhao, and Liu}]{song-etal-2018-structure}
Kaiqiang Song, Lin Zhao, and Fei Liu. 2018.
\newblock \href {https://www.aclweb.org/anthology/C18-1146} {Structure-infused
  copy mechanisms for abstractive summarization}.
\newblock In \emph{Proceedings of the 27th International Conference on
  Computational Linguistics}, pages 1717--1729, Santa Fe, New Mexico, USA.
  Association for Computational Linguistics.

\bibitem[{Strakov{\'a} et~al.(2019)Strakov{\'a}, Straka, and
  Haji{\v{c}}}]{strakova2019neural}
Jana Strakov{\'a}, Milan Straka, and Jan Haji{\v{c}}. 2019.
\newblock Neural architectures for nested ner through linearization.
\newblock \emph{arXiv preprint arXiv:1908.06926}.

\bibitem[{Tenney et~al.(2019)Tenney, Das, and Pavlick}]{tenney2019bert}
Ian Tenney, Dipanjan Das, and Ellie Pavlick. 2019.
\newblock Bert rediscovers the classical nlp pipeline.
\newblock \emph{arXiv preprint arXiv:1905.05950}.

\bibitem[{Vinyals et~al.(2015)Vinyals, Fortunato, and
  Jaitly}]{vinyals2015pointer}
Oriol Vinyals, Meire Fortunato, and Navdeep Jaitly. 2015.
\newblock \href {http://arxiv.org/abs/1506.03134} {Pointer networks}.

\bibitem[{Wang and Lu(2018)}]{wang-lu-2018-neural}
Bailin Wang and Wei Lu. 2018.
\newblock \href {https://doi.org/10.18653/v1/D18-1019} {Neural segmental
  hypergraphs for overlapping mention recognition}.
\newblock In \emph{Proceedings of the 2018 Conference on Empirical Methods in
  Natural Language Processing}, pages 204--214, Brussels, Belgium. Association
  for Computational Linguistics.

\bibitem[{Wang et~al.(2018)Wang, Lu, Wang, and
  Jin}]{wang-etal-2018-neural-transition}
Bailin Wang, Wei Lu, Yu~Wang, and Hongxia Jin. 2018.
\newblock \href {https://doi.org/10.18653/v1/D18-1124} {A neural
  transition-based model for nested mention recognition}.
\newblock In \emph{Proceedings of the 2018 Conference on Empirical Methods in
  Natural Language Processing}, pages 1011--1017, Brussels, Belgium.
  Association for Computational Linguistics.

\bibitem[{Yang et~al.(2019)Yang, Dai, Yang, Carbonell, Salakhutdinov, and
  Le}]{Yang19XLNet}
Zhilin Yang, Zihang Dai, Yiming Yang, Jaime Carbonell, Russ~R Salakhutdinov,
  and Quoc~V Le. 2019.
\newblock \href
  {http://papers.nips.cc/paper/8812-xlnet-generalized-autoregressive-pretraining-for-language-understanding.pdf}
  {Xlnet: Generalized autoregressive pretraining for language understanding}.
\newblock In \emph{Advances in Neural Information Processing Systems 32}, pages
  5753--5763. Curran Associates, Inc.

\bibitem[{Zhang et~al.(2019)Zhang, Ma, Duh, and
  Van~Durme}]{zhang-etal-2019-amr}
Sheng Zhang, Xutai Ma, Kevin Duh, and Benjamin Van~Durme. 2019.
\newblock \href {https://doi.org/10.18653/v1/P19-1009} {{AMR} parsing as
  sequence-to-graph transduction}.
\newblock In \emph{Proceedings of the 57th Annual Meeting of the Association
  for Computational Linguistics}, pages 80--94, Florence, Italy. Association
  for Computational Linguistics.

\bibitem[{Zhou et~al.(2018)Zhou, Yang, Wei, and Zhou}]{zhou2018sequential}
Qingyu Zhou, Nan Yang, Furu Wei, and Ming Zhou. 2018.
\newblock Sequential copying networks.
\newblock In \emph{Thirty-Second AAAI Conference on Artificial Intelligence}.

\end{thebibliography}
\bibliographystyle{acl_natbib}

\clearpage
\appendix
\section{Appendices}
\label{sec:appendix}

\subsection{Hyper-parameter searching}
\label{sec:app_expr}

We manually tune the hyper-parameters according to the performance of the model, i.e. the dev F1 scores. 
The hyper-parameters include the number of the stacked BiLSTM layers,
the number of RoBERTa layers,
and the subword pooling (use it or not). 
For the encoders or word embeddings, we used GloVe \cite{pennington2014glove}, ELMo \cite{peters-etal-2018-deep}, SpanBERT \cite{joshi2019spanbert}, XLNet \cite{Yang19XLNet}, BERT \cite{devlin-etal-2019-bert}, and RoBERTa \cite{roberta}.
After picking RoBERTa, we tried 1, 2, and 3 layers of stacked LSTM layers,
and 1, 5, 10, 15, 20, 24 layers of RoBERTa.
Among these trials, the model we adopt is with 2 layers of BiLSTM encoder and decoder, 24 layers of RoBERTa, and no subword pooling.
Finally, we also benchmarked decoding strategy with beam search. With beam size 10, we gained 0.4 F1 over our greedy model but almost 70x slower. We leave \textit{efficient} decoding strategies to future work.

The searching procedure and the intermediate results are shown in \autoref{table:subwordpooling}, \autoref{table:bilstmlayers}, and \autoref{table:robertalayers}.

\begin{table}[ht]
\centering
\begin{tabular}{lrrr}
 \toprule
  & P     & R     & F1    \\ \midrule
Subword pooling & 89.5 & 79.5 & 84.2  \\ 
No pooling      & 89.6 & 84.5 & 87.0  \\ \bottomrule
\end{tabular}
\caption{Comparison between using subword embeddings generated by RoBERTa directly, or pooling them into tokens representations, as evaluated on the NNE development set.}
\label{table:subwordpooling}
\end{table}

\begin{table}[ht]
    \centering
    \begin{tabular}{crrr} 
    \toprule
     Number of layers       & P     & R     & F1    \\ \midrule
    1  & 90.6 & 82.6 & 86.4  \\ 
    2  & 89.6   & 84.5 & 87.0   \\ 
    3  & 87.8 & 83.1 & 85.4  \\
    \bottomrule
\end{tabular}
\caption{Performance of models with different numbers of layers in stacked BiLSTM encoder and decoder (on NNE development set).}
\label{table:bilstmlayers}
\end{table}

\begin{table}[ht]
\centering
\begin{tabular}{rrrrr}\toprule
layer   & P     & R     & F1    \\ \midrule
1        & 86.4 & 73.9 & 79.7                                      \\ 
5        & 89.0 & 80.2 & 84.4                                      \\ 
10       & 90.9 & 82.3 & 86.4                                      \\ 
15       & 91.3 & 83.2 & 87.1                                      \\ 
20       & 91.0 & 81.0 & 85.7                                      \\ 
24 (last)& 90.6 & 82.6 & 86.4                                     \\  \bottomrule                                     
\end{tabular}
\caption{Performance on the NNE development set using different layers of RoBERTa large as the input representation.}
\label{table:robertalayers}
\end{table}

\subsection{Data}

The nested named entity dataset is available online at \url{https://github.com/nickyringland/nested_named_entities}. Following \citet{ringland-etal-2019-nne}, we use section 02 for development (1,989 sentences), sections 23 and 24 for testing (3,762 sentences), and the remaining sections for training (43,457 sentences).

\subsection{Example of linearization}

\autoref{fig:linear_example} highlights another example of our linearization strategy. In our final strategy, ties are broken randomly when spans have multiple labels (such as ``Smith Barney'') in the example. We did try sorting those spans by some deterministic method, such as label frequency (in the training corpus). We found that deterministically sorting these did not improve performance, sometimes even hurting.

\definecolor{clover}{RGB}{17,128,2}
\definecolor{cayenne}{RGB}{128,0,2}
\definecolor{eggplant}{RGB}{64,0,128}

\begin{table*}[ht]
    \begin{tabular}{|c|}
    \hline
    $\underset{\color{black}{\text{PER}}}{\colorbox{Plum4}{\text{$\underset{\color{black}{\text{FIRST}}}{\colorbox{Salmon1}{\text{James}}}$ $\underset{\color{black}{\text{NAME}}}{\colorbox{LightSteelBlue1}{\text{Wilbur}}}$}}}$, a  
    
    $\overset{\color{black}{\text{ORGCORP}}}{{\text{$\underset{\color{black}{\text{NAME}}}{\colorbox{LightSkyBlue3}{\text{$\underset{\color{black}{\text{NAME}}}{\colorbox{LightSteelBlue1}{\text{Smith}}}$ $\underset{\color{black}{\text{NAME}}}{\colorbox{LightSteelBlue1}{\text{Barney}}}$}}}$}}}$
    
     analyst   \\ \hline
     \textcolor{cayenne}{0} \textcolor{clover}{CN} \textcolor{eggplant}{PER} \textcolor{cayenne}{0} \textcolor{eggplant}{FIRST} \textcolor{cayenne}{1} \textcolor{eggplant}{NAME} \textcolor{cayenne}{4} \textcolor{clover}{CN} \textcolor{eggplant}{NAME} \textcolor{cayenne}{4} \textcolor{clover}{CN} \textcolor{eggplant}{ORGCORP} \textcolor{cayenne}{4} \textcolor{eggplant}{NAME} \textcolor{cayenne}{5} \textcolor{eggplant}{NAME}\\ \hline

    \end{tabular}
\caption{Example of linearization of a structured output of nested named entities. Spans are in ascending order of starting index and ties are broken by span length.}
\label{fig:linear_example}
\end{table*}

\subsection{Examples of errors}
\label{sec:app_err}
We present several examples of errors in \autoref{table:errors}. There are four major types. Two types are partially correct: (1) correct span boundary prediction but incorrect label; (2) incorrect span boundaries (still overlaps heavily with the correct span) but with the correct label. The other two types are (3) incorrect span and label, which combines both of the above errors, and (4) missing span entirely. Error types (2), (3), and (4) all affect recall (specifically span recall) and could provide further insight on how to improve our model's recall. We did not find many instances where spurious spans are predicted.

\begin{table*}
\begin{tabularx}{\textwidth}{ lX}
\toprule
\multicolumn{2}{l}{\textbf{Correct span, incorrect label}}                                             \\ \midrule
\emph{\small{predict}} & \small{... to be the case in $\overset{\color{black}{\text{COUNTRY}}}{\colorbox{Peru}{\text{Sing apore}}}$ , a country of about three million people with a rel atively high soft - dr ink cons umption rate – a key ind icator of $\underset{\color{black}{\text{NAME}}}{\colorbox{PaleTurquoise}{\text{C oke}}}$ ’s success in a market .} \\ \midrule
\emph{\small{gold}}  & \small{... to be the case in $\overset{\color{black}{\text{CITYSTATE}}}{\colorbox{LavenderBlush2}{\text{Sing apore}}}$ , a country of about three million people with a rel atively high soft - dr ink cons umption rate – a key ind icator of $\underset{\color{black}{\text{ORGCORP}}}{\colorbox{LightGoldenrod1}{\text{C oke}}}$ ’s success in a market .}  \\ \midrule
\multicolumn{2}{l}{\textbf{Incorrect span, correct label}} \\ \midrule
\emph{\small{predict}} &  \small{“ Nothing can be better than this , ” s ays Don S ider , owner of the $\underset{\color{black}{\text{CITY}}}{\colorbox{LightSalmon1}{\text{West Pal m}}}$ Be ach T rop ics .}  \\ \midrule
\emph{\small{gold}} &  \small{“ Nothing can be better than this , ” s ays Don S ider , owner of the $\underset{\color{black}{\text{CITY}}}{\colorbox{LightSalmon1}{\text{West Pal m Be ach}}}$ T rop ics .}  \\ \midrule
\multicolumn{2}{l}{\textbf{Incorrect span and label}}                                   \\ \midrule
\emph{\small{predict}} & \small{Pro ct er Gam ble Co . recent ly introdu ced ref ill able versions of four products including T ide and Mr . $\overset{\color{black}{\text{NAME}}}{\colorbox{LightSteelBlue1}{\text{Clean}}}$ , in Canada , but doesn ’t plan to bring them to the U . S . .}\\ \midrule
\emph{\small{gold}}  &   \small{Pro ct er Gam ble Co . recent ly introdu ced ref ill able versions of four products including T ide and $\overset{\color{black}{\text{ANIMATE}}}{\colorbox{Goldenrod}{\text{Mr . Clean}}}$ , in Canada , but doesn ’t plan to bring them to the U . S . .}\\ \midrule
\multicolumn{2}{l}{\textbf{Missing entities}}    \\ \midrule
\emph{\small{predict}}  & \small{C ERT IFIC ATES OF DE POS IT : 8 . 09 \% one month ; 8 . 04 \% two months ; 8 . 03 \% three months ; 7 . 96 \% six months ; 7 . 92 \% one year .} \\ \midrule
\emph{\small{gold}}  &  \small{C ERT IFIC ATES OF DE POS IT : 8 . 09 \% $\overset{\color{black}{\text{CARDINAL}}}{\colorbox{DarkSeaGreen4}{\text{one}}}$ month ; $\overset{\color{black}{\text{PERCENT}}}{\colorbox{DarkSeaGreen1}{\text{$\underset{\color{black}{\text{CARDINAL}}}{\colorbox{DarkSeaGreen4}{\text{8 . 04}}}$ \%}}}$ two months ; 8 . 03 $\overset{\color{black}{\text{UNIT}}}{\colorbox{Pink2}{\text{\%}}}$ three months ; 7 . 96 \% six months ; 7 . 92 \% one year .} \\ \bottomrule

\end{tabularx}
\caption{Different types of error made by the CopyNext model in the NNER task.}
\label{table:errors}
\end{table*}

\end{document}